\documentclass{article} % For LaTeX2e
\usepackage{iclr2016_conference,times}
\usepackage{mathptmx}
\usepackage{url,latexsym,amsmath,amsthm,xspace,rotating,multirow,multicol,xspace,amssymb,paralist}
\usepackage{euscript}
\usepackage{wrapfig}
\usepackage{fancybox,xcolor}
\usepackage{longtable}
\usepackage{paralist}
\usepackage{subfigure}
\usepackage[normalem]{ulem}
\usepackage[pdftex]{hyperref}
\usepackage{alphalph}

\usepackage{url}
\usepackage{latexsym}

\usepackage{amsmath}
\usepackage{amsthm}
\usepackage{amssymb}
\usepackage{graphicx}
\usepackage{xspace}
\usepackage{tabularx}
\usepackage{multicol}
\usepackage{multirow}
\usepackage{url}
\usepackage{wrapfig}
\usepackage{comment}
\usepackage{listings}
\usepackage{color}
\usepackage[utf8]{inputenc}
\usepackage{fancyvrb}
\usepackage{booktabs}
\usepackage{color}
\usepackage[normalem]{ulem}

\newcommand{\bmx}[0]{\begin{bmatrix}}
\newcommand{\emx}[0]{\end{bmatrix}}

\newcommand{\vect}[1]{\mathbf{#1}}
\newcommand{\vects}[1]{\boldsymbol{#1}}
\newcommand{\matr}[1]{\mathbf{#1}}

\newcommand{\vh}[0]{\vect{h}}

\newcommand{\vx}[0]{\vect{x}}
\newcommand{\vz}[0]{\vect{z}}

\newcommand{\mW}[0]{\matr{W}}

\newcommand{\mC}{\matr{C}}

\newcommand{\TT}[0]{\vects{\theta}}

\newcommand{\vmu}[0]{\vects{\mu}}

\newcommand{\NN}[0]{\mathcal{N}}
\newcommand{\LL}[0]{\mathcal{L}}

\newcommand{\E}[0]{\mathbb{E}}

\newcommand{\IOU}{\text{IOU}}
\newcommand{\fout}{\text{out}}
\newcommand{\conv}{\text{conv}}
\newcommand{\rec}{\text{rec}}

\title{First Step toward \\ Model-Free, Anonymous Object Tracking \\ with
Recurrent Neural Networks}

\author{Quan Gan$^\star$$^\diamond$ 
\And
Qipeng Guo$^\star$$^\diamond$ 
\And
Zheng Zhang$^\bullet$$^\ast$ 
\And
Kyunghyun Cho$^\bullet$$^\circ$ 
}

\begin{document}

\let\thefootnote\relax\footnotetext{
\hspace{-2.25em}
$\star$ Equal contribution \\
$\diamond$ 
Fudan University, Shanghai  \\
\texttt{qgan11@fudan.edu.cn,guoqipeng831@gmail.com} \\
$\bullet$
Courant Institute of Mathematical Sciences, 
New York University 
\texttt{\{zz,kyunghyun.cho\}@nyu.edu} \\
$\ast$
NYU Shanghai \\
$\circ$
Center for Data Science, New York University 
}

\maketitle

\begin{abstract}

    In this paper, we propose and study a novel visual object tracking approach
    based on convolutional networks and recurrent networks. The proposed
    approach is distinct from the existing approaches to visual object tracking,
    such as filtering-based ones and tracking-by-detection ones, in the sense
    that the tracking system is explicitly trained off-line to track {\em
    anonymous} objects in a noisy environment. The proposed visual tracking
    model is end-to-end trainable, minimizing any adversarial effect from
    mismatches in object representation and between the true underlying dynamics
    and learning dynamics. We empirically show that the proposed tracking
    approach works well in various scenarios by generating artificial video
    sequences with varying conditions; the number of objects, amount of noise
    and the match between the training shapes and test shapes. 

\end{abstract}

\section{Introduction}

Visual object tracking is a problem of building a computational model that is
able to predict the location of a designated object from a video clip consisting
of many consecutive video frames. Using deep learning techniques, we attack this
problem with a model-free statistical approach.

Our goal in this paper is to build a model that can track an {\em anonymous}
object in an image sequence. This task finds immediate applications in important
scenarios such as self-driving cars. As safety is first-order consideration,
identifying what class the object belongs to is much less critical than
identifying their whereabouts to avoid a collision. It is also an important step
towards dramatically improving the generalizability of a tracking system, since
real-world objects far exceeds labelled categories.

Our model integrates convolutional network with recurrent network, and deploys
attention at multiple representation layers. The recurrent network outputs a
bounding box prediction of the target object. It fuses past predictions along
with their corresponding visual features produced by the convolutional network.
Finally, the prediction can optionally emphasizes attention areas in the input
before feeding it into convolutional network. The entire model is end-to-end
trained off-line. We use synthesized data set that simulates changing trajectory
and acceleration of the target, various degree of foreground occlusions, and
distraction of background clutter and other targets. Experimental results show
that our model delivers robust performance.

%As we will show empirically later in this paper, a model-free deep learning
%framework adds itself to the rich literature of object tracking research.
%However, it also aims at indirectly improving the capability of deep learning
%systems themselves.  The success of deep learning largely depends on the amount
%of labelled data.  Videos contain far more objects of high variations, and is
%thus an important data source.  Yet, unlike still image, labelling correctly on
%objects in the video can be exceedingly expensive. As an example, one of the
%largest video dataset contains over one million Youtube videos, but all of them
%only have video-level labels \citep{KarpathyCVPR14}.  The ability to track
%anonymous objects provides weak labels by segmenting them out.  If successful, a
%single class label of any one frame can propagate to the rest of the frames.
%Thus, tracking anonymous objects provides a foundation for scalable,
%semi-automatic labelling, unlocking the rich data source currently trapped in
%unlabelled videos.

The rest of the paper is organized as follows. We start by reviewing two
important categories of conventional visual object tracking in
Sec.~\ref{sec:background}, which are filtering-based tracking and
tracking-by-detection. In Sec.~\ref{sec:model}, we describe a novel recurrent
tracking model we propose in this paper, followed by discussing related works in
Sec.~\ref{sec:related}. The settings for experiments are extensively described
in Sec.~\ref{sec:data}--\ref{sec:model_train}, which is followed by the results
and analysis in Sec.~\ref{sec:result}. We finalize this article with potential
future research directions in Sec.~\ref{sec:conclusion}.

\section{Background: Visual Object Tracking and Limitations}
\label{sec:background}

A system for visual object tracking often comprises two main components; object
detection and object tracking. Object detection refers to the process by which a
designated object in each video frame is detected, while object tracking refers
to the process of continually predicting the location of the detected object.

The goal of {\em object detection} is to extract a feature vector $\phi(x)$ of
an object which often encodes both the object's shape and location, given each
video frame $x$. The specifics of the feature vector heavily depend on the
choice of {\em object representation}. For instance, if an object is represented
as a point, the feature vector of the detected object is a two-dimensional
coordinate vector corresponding to the center of gravity of the object in each
frame. 

Object detection is followed by {\em object tracking}. There are many approaches
proposed over a number of decades (see, e.g., \cite{yilmaz2006object},) and we
are interested in this proposal a statistical approach. A statistical approach
assumes that the feature vector $\phi(x)$ from the object detection stage is a
noisy observation of the true, underlying state (location) of the object which
is not observed. The goal is to infer the underlying state for each video frame,
given a sequence of observations, i.e., features vectors.

\paragraph{Filtering-based Visual Object Tracking}

In filtering-based object tracking, it is natural to establish a probabilistic
graphical model, or often referred to as a state-space model, by defining
\begin{itemize}
    \itemsep -.3em
    \item Observation model: $p(\phi(\vx_t) | \vz_t)$
    \item Transition model: $p(\vz_t | \vz_{t-1})$,
\end{itemize}
where $\phi(\vx_t)$ and $\vz_t$ are the observation and the latent state at time
$t$. 

One of the most well known filtering-based tracking model is so called {\em
Kalman filter} \citep{broida1986estimation}. Kalman filter assume that both the
observation and transition models are Gaussian distributions such that
\begin{align*}
    \phi(\vx_t)|\vz_t \sim \NN(\mW_x \vz_t, \mC_x), &
    ~~~\vz_t | \vz_{t-1} \sim \NN(\mW_z \vz_{t-1}, \mC_z),
\end{align*}
where $\NN(\vmu, \mC)$ is a Gaussian distribution with its mean $\vmu$ and
covariance matrix $\mC$. 

\begin{figure}[t]
    \centering
    \begin{minipage}{0.49\textwidth}
        \centering
        \includegraphics[width=0.99\columnwidth]{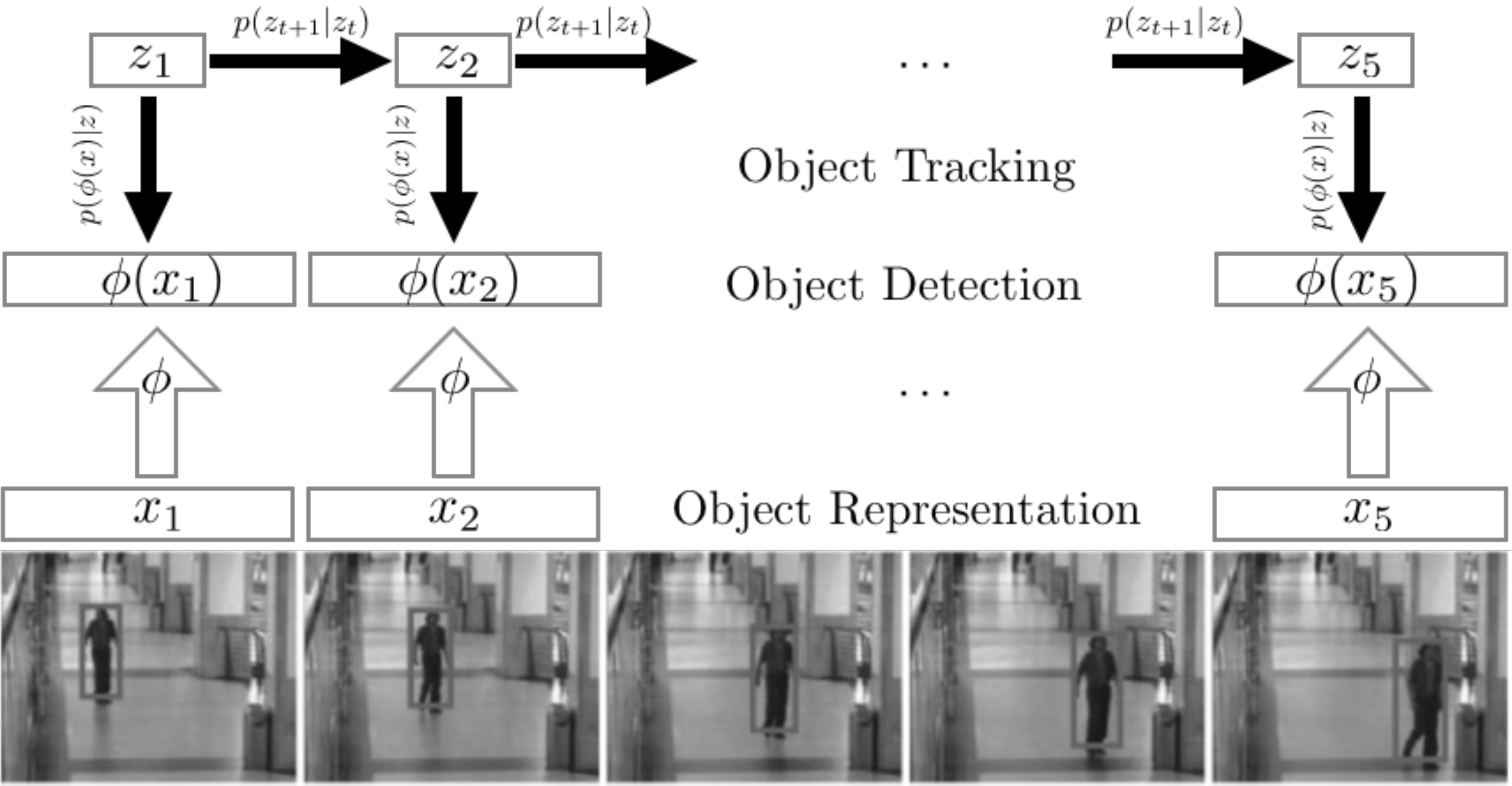}
    \end{minipage}
    \hfill
    \begin{minipage}{0.49\textwidth}
        \centering
        \includegraphics[width=0.99\columnwidth]{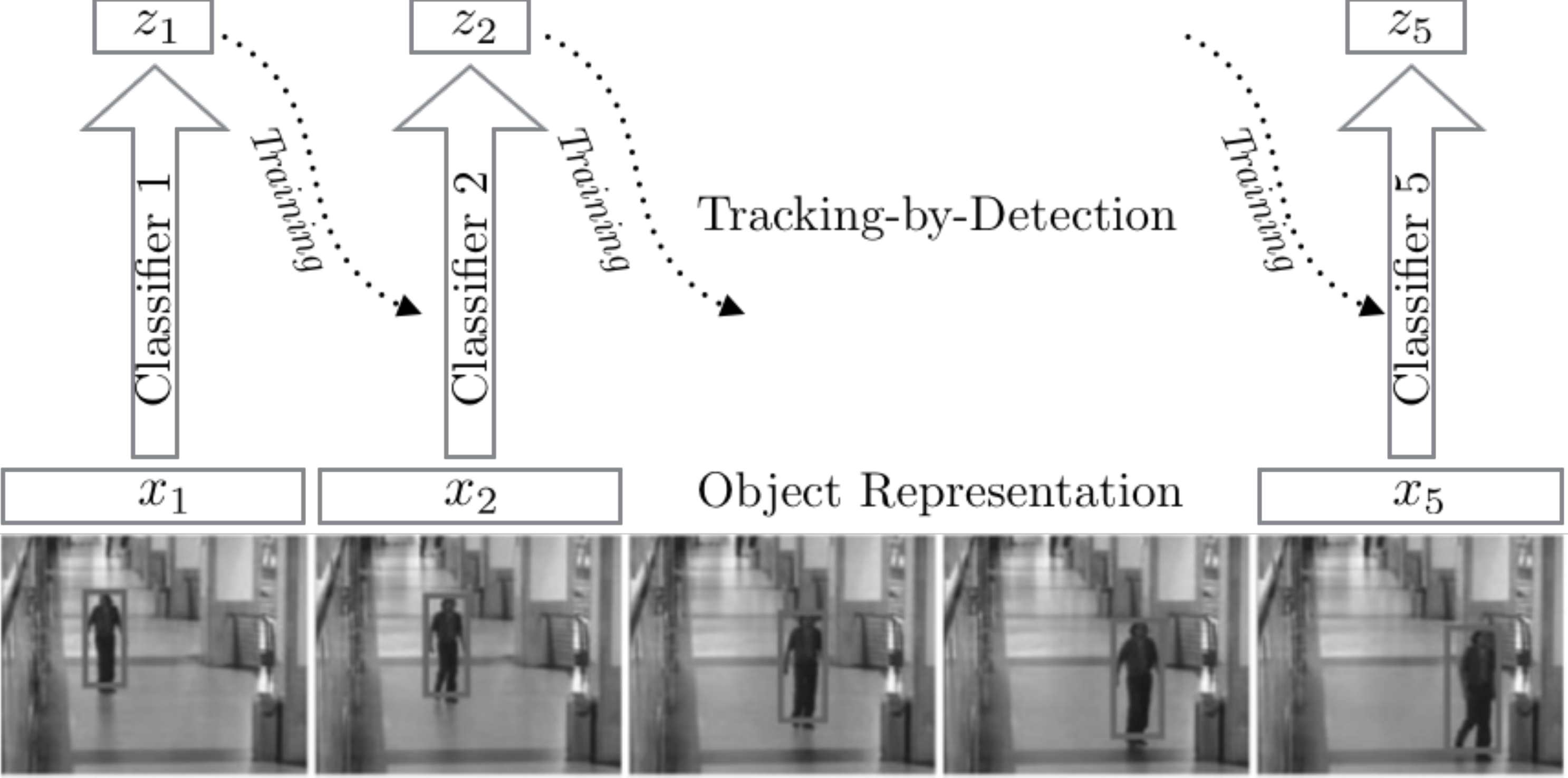}
    \end{minipage}

    \begin{minipage}{0.49\textwidth}
        \centering
        (a) Filtering-based Tracking
    \end{minipage}
    \hfill
    \begin{minipage}{0.49\textwidth}
        \centering
        (b) Tracking-by-Detection
    \end{minipage}

    \caption{
        Graphical illustration of the filtering-based tracking and
        tracking-by-detection.
    }
    \label{fig:conventional}
\end{figure}

Once the model is defined, the goal is to infer the following posterior
distribution:
\begin{align}
    \label{eq:posterior}
    p(\vz_t | \phi(x_1), \phi(x_2), \ldots \phi(\vx_t)).
\end{align}
In other words, we try to find the distribution over the potential object
location in the $t$-th video frame given all the detected objects (represented
by the feature vectors) up to the $(t-1)$-th frame. See
Fig.~\ref{fig:conventional} (a) for the graphical illustration.

In this scheme, object detection and tracking are largely considered separate
from each other. This means that the object tracking stage is designed to work
while being blind to the object representation or feature vectors from the
object detection stage. However, this is not to say that there is no effect of
the choice of object and/or feature representation, as mismatch between the
model definition in object tracking and the distribution based on the selected
feature representation will lead to suboptimal tracking result.

\paragraph{Tracking-by-Detection}

An approach more relevant to our proposal is {\em tracking-by-detection} (see,
e.g., \cite{li2013survey}.) This approach is more holistic than the previous
approach, because a single model is trained online to track an object in
interest. In other words, tracking-by-detection builds a discriminative model
that directly approximates the posterior distribution over the underlying
location/state of the object in Eq.~\eqref{eq:posterior}. See
Fig.~\ref{fig:conventional} (b) for the graphical illustration.

Often, tracking-by-detection is done not as regression but as classification of
regions of a video frame. The classifier is initialized to work well on the
first few frames where a separate object detector or human expert has classified
each region of the frames as either foreground or background. This classifier is
used to detect/track the object in the next frame of which ground-truth labeling
is not available. Once each region in the next frame is classified, the
classifier is further {\em fine-tuned} with these new examples. This continues
until the video clip reaches its end or the object disappears (all regions are
classified negative.)

This approach of tracking-by-detection has a number of limitations, of which the
most severe ones are drifting (accumulation of error over multiple frames) and
inability to handle occlusion easily. We notice that both of these issues arise
from the fact that this approach effectively assumes Markov property over the
posterior distribution in Eq.~\eqref{eq:posterior}, meaning
\begin{align*}
    p(\vz_t | \phi(\vx_1), \phi(\vx_2), \ldots \phi(\vx_t))) \approx
    p(\vz_t | \phi(\vx_{t-1})).
\end{align*}

Because the tracking model considers only two consecutive frames (or more
precisely concentrates heavily on a latest pair of consecutive frames only), it
is not possible for the model to adjust for accumulated error over many
subsequent frames. If the model has access to the whole history of frames and
its own prediction of the object's locations, it will be possible for the model
to take the errors made throughout the video clip and address the issue of
drifting.

The tracking model's lack of access to the history of the previous frames makes
it nearly impossible for the model to handle occlusion. On the other hand, if
the tracking model has the history of its previous observations and predictions
of the object's location, it can more easily infer that the object has not
disappeared totally from the view but moved behind some other background
objects. The tracking model may even be able to infer the object's location even
while it is being hidden from the view by understanding its motion based on the
history the object's locations.

\paragraph{Limitations}

We have noticed three limitations in these conventional approaches. First, {\em
object representation is designed independently} from the rest of the tracking
pipeline, and any suboptimal choice of object representation may negatively
impact the subsequent stages.  Second, the filtering-based approach is {\em not
robust to a mismatch between the underlying model description and the reality}.
Third, {\em the lack of access to the history} of all the previous video frames
makes tracking-by-detection sensitive to complex motions and structured noise
such as occlusion. Lastly, tracking-by-detection requires {\em a classifier, or
a regressor, to be tuned at each frame}, making it less suitable to be applied
in real-time applications.

\section{Visual Object Tracking with Deep Neural Networks}
\label{sec:model}

Here we describe a novel approach to visual anonymous object tracking by using
techniques from deep learning~\citep{lecun2015deep}. Our aim in introducing a
novel approach is to build a system that avoids the four limitations of the
conventional visual tracking systems we discussed in the earlier section.

%\begin{figure}[t]
\begin{wrapfigure}{R}{0.49\textwidth}
    \centering
    \includegraphics[width=0.49\columnwidth]{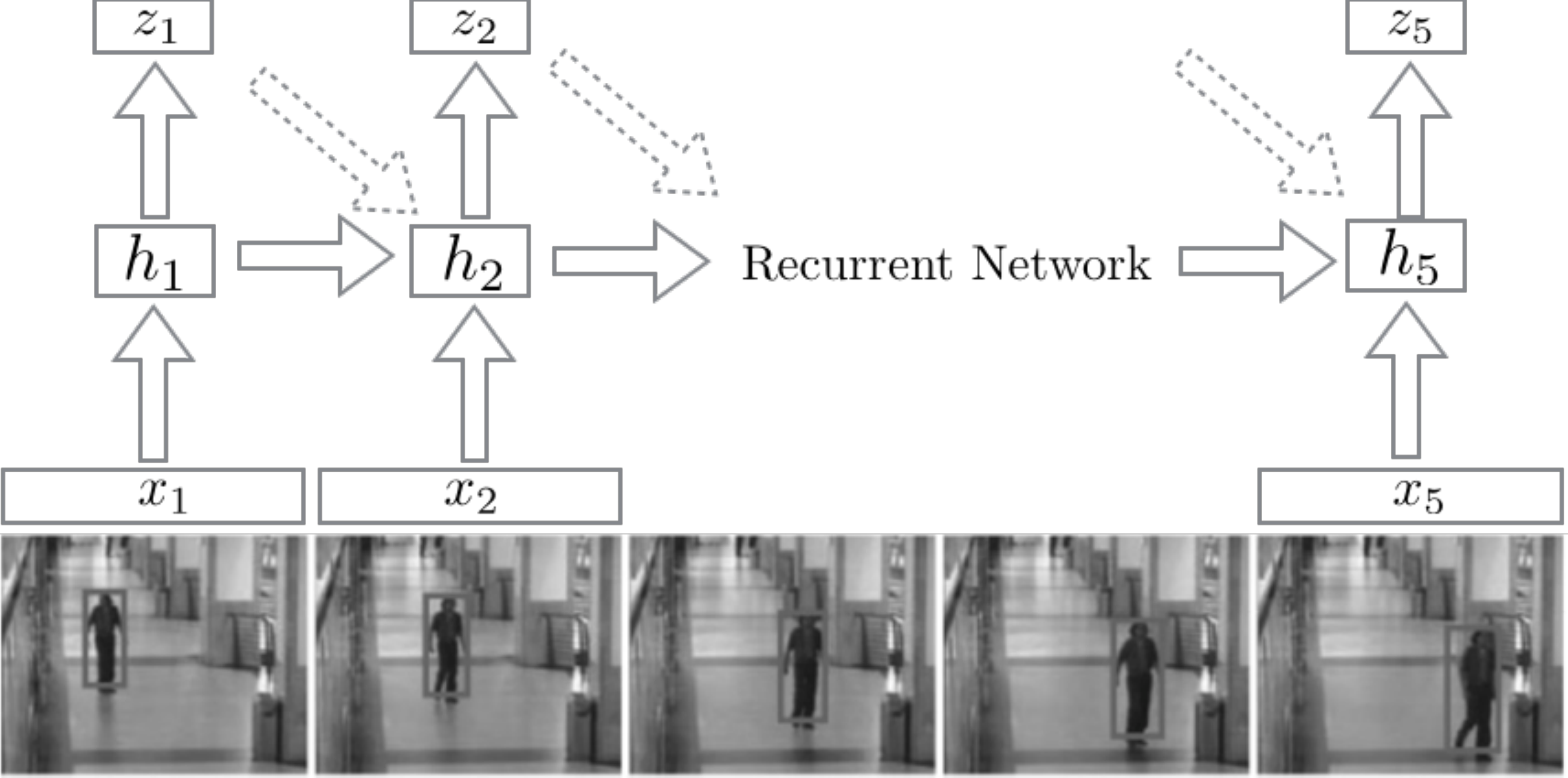}
    \caption{
        Graphical illustration of the visual tracking system proposed here.
    }
    \label{fig:proposed}
\end{wrapfigure}
%\end{figure}

The proposed model is a deep recurrent network that takes as input raw video
frames of pixel intensities and returns the coordinates of a bounding box of an
object being tracked for each frame. Mathematically, this is equivalent to
saying that the proposed model factorizes the full tracking probability into
\begin{align*}
    p(\vz_1, \vz_2, \ldots, \vz_T | \vx_1, \vx_2, \ldots, \vx_T) 
    = \prod_{t=1}^T p(\vz_t | \vz_{<t}, \vx_{\leq t}),
\end{align*}
where $\vz_t$ and $\vx_t$ are the location of an object and an input frame,
respectively, at time $t$. $\vz_{<t}$ is a history of all previous locations
before time $t$, and $\vx_{\leq t}$ is a history of all input frames up to time
$t$.

\subsection{Model Description}
\label{sec:description}

At each time step $t$, an input frame $\vx_t$ is first processed by a
convolutional network, which has recently become a {\em de facto} standard in
handling visual input \citep[see,
e.g.,][]{lecun1998efficient,krizhevsky2012imagenet}. This results in a feature
vector $\phi(\vx_t)$ of the input frame $\vx_t$:
\begin{align}
    \label{eq:rnn_detect}
    \phi(\vx_t) = \conv_{\TT_c} (m(\vx_t, \tilde{\vz}_{t-1})),
\end{align}
where $\conv_{\TT_c}(\cdot)$ is a convolutional network with its parameters
$\TT_c$, and $m(\cdot, \cdot)$ is a preprocessing routine for the raw frame. We will
discuss this preprocessing routine at the end of this section.
$\tilde{\vz}_{t-1}$ is the predicted location of an object from the previous
frame $\vx_{t-1}$, which we describe in more detail below. 

This feature vector of the input frame is fed into a recurrent neural network.
The recurrent neural network updates its internal memory vector $\vh_t$ based on
the previous memory vector $\vh_{t-1}$, previous location of an object
$\tilde{\vz}_{t-1}$ and the current frame $\phi(\vx_t)$:
\begin{align}
    \label{eq:rnn_update}
    \vh_t = \rec_{\TT_r}(\vh_{t-1}, \tilde{\vz}_{t-1}, \phi(\vx_t)),
\end{align}
where $\rec_{\TT_r}$ is a recurrent activation function such as gated recurrent
units~\cite{cho2014learning}, long short-term memory
units~\cite{hochreiter1997long} or a simple logistic function, parametrized with
the parameters $\TT_r$. This formulation lets the recurrent neural network to
summarize the history of predicted locations $\vz_{<t}$ and input frames
$\vx_{\leq t}$ up to time step $t$.

With the newly updated memory state $\vh_t$, the recurrent neural network
computes the predictive distribution over the object's location (see
Eq.~\eqref{eq:posterior}. This is done
again by a deep neural network $\fout_{\TT_o}$ \cite{pascanu2014construct}:
\begin{align}
    \label{eq:rnn_predictive_dist}
    p(\vz_t | \vz_{<t}, \vx_{\leq t}) = \fout_{\TT_o}(\vh_t),
\end{align}
where $\TT_o$ is a set of parameters defining the output neural network. We take
the mean of this predictive distribution as a predicted location $\tilde{\vz}_t$
at time $t$:
\begin{align}
    \label{eq:rnn_predict}
    \tilde{\vz}_t = \E\left[ \vz | \vz_{<t}, \vx_{\leq t}  \right].
\end{align}

This whole process (Eqs.~\eqref{eq:rnn_detect}--\eqref{eq:rnn_predict}) is
iteratively applied as a stream of new frames arrives. This is graphically
illustrated in Fig.~\ref{fig:proposed}.

\paragraph{Preprocessing Input Frame $\vx_t$}

We considered a number of possible strategies for building a preprocessing
routine $m(\cdot, \cdot)$ from Eq.~\eqref{eq:rnn_detect}. The most obvious and
straightforward choice is to simply have an identity function, 
%which will be
equivalent to simply passing a raw frame into the convolutional network
$\conv_{\TT_c}$. In this case, we use an identity
function to preprocess each input frame $\vx_t$:
\begin{align}
\label{eq:preprocess_id}
m(\vx_t, \tilde{\vz}_{t-1}) = \vx_t,
\end{align}
where $\tilde{\vz}_{t-1}$ is ignored. This is equivalent to simply letting the
tracker have a full, unadjusted view of the whole frame.

% we will spell it out in later section anyways....
%\begin{align*}
%    m(\vx_t, \vz_t) = \vx_t
%\end{align*}

On the other hand, we can design a preprocessing function $m$ such that it will
facilitate tracking. One possible choice is to weight each pixel of the raw
frame $\vx_t$ such that a region surrounding the predicted location of an object
in the previous frame is given higher weights. This will help the subsequent
layers, i.e., $\conv_{\TT_c}$, $\rec_{\TT_r}$ and $\fout_{\TT_o}$, to focus more
on that region. We refer to this preprocessing routine as an {\em attentive weight scheme}.

In this model, the recurrent network outputs the coordinates of (1) top-left
corner $(x_0, y_0)$, (2) bottom-right corner $(x_1, y_1)$, (3) log-scale $s$,
(4) log-ratio $r$ between the stride and the image size, and (5) log-amplitude
$a$. Given these output, we weight each pixel using a mixture of $N \times N$
Gaussians. Each Gaussian $(i,j)$ is centered at
\[
\left(
\frac{x_0 + x_1}{2} + \left( i - \frac{N}{2} - 0.5 \right) \exp\left( r \right) K, 
\frac{y_0 + y_1}{2} + \left( j - \frac{N}{2} - 0.5 \right) \exp\left( r \right) K 
\right),
\]
and has the standard deviation of $\exp\left(s\right) \frac{K}{2}$. $K$
corresponds to the width or height of a frame. These Gaussians are used to form
a mask $G(\vz_{t-1})$ which is used as 
\begin{align}
\label{eq:preprocess_att}
m(\vx_t, \vz_{t-1}) = \vx_t \cdot G(\vz_{t-1})
%    {\color{red} \text{Andy, please fill here}}
\end{align}

We emphasize that
we only weight the pixels of each frame and do {\em not} extract a patch, as was
done by \citet{gregor2015draw} and \citet{kahou2015ratm}, because this approach
of ignoring an out-of-patch region may lose too much information needed for
tracking.

\subsection{Training}

Unlike the existing approaches to visual object tracking, described in
Sec.~\ref{sec:background}, we take an off-line training strategy. This means
that the trained model is used as is in test time. 

An obvious advantage to this approach is that there is no overhead in finetuning
the model on-the-fly during test time. On the other hand, there is an obvious
disadvantage that the model must be able to track an object whose shape or
texture information was {\em not} available during training time.  We propose
here a training strategy that aims at overcoming this disadvantage.
% of the pretrained tracking system.

We were motivated from recent observations from many research groups that a deep
convolutional network, pretrained on a large image dataset of generic, natural
images, is useful for subsequent vision tasks which may not necessarily involve
the same types of objects~\citep[see,
e.g.,][]{sermanet2013overfeat,bar2015deep}. As the model we propose here
consists of a convolutional network and recurrent network, we expect a similar
benefit by training the whole model with generic shapes, which may not appear
during test time. 

As usual when training a deep neural network, we use stochastic gradient descent
(SGD). At each SGD update, we {\em generate} a minibatch of examples by the
following steps:
\begin{enumerate}
    \itemsep -.3em
    \item Select a random background image from a large set of image.\footnote{
            We can use, for instance, ImageNet Data from
            \url{http://www.image-net.org/}, or create one with random clutters
            as in MNIST cluttered from
            \url{https://github.com/deepmind/mnist-cluttered}.
        }
    \item Randomly choose a shape of an object from a predefined set of generic
        shapes.
    \item Create a sequence of frames by randomly moving the selected object with cluttered background and foreground.
%        with the static background.
    \item (Optional) Add various types of noise, including motion and scale change of both object and clutters.
\end{enumerate}
After these steps, each training example is a pair of a video clip, which
contains a randomly chosen background and a moving shape, and a sequence of
ground-truth locations, i.e., $((\vx_1, \vz_1^*), \ldots, (\vx_T, \vz_T^*))$. 

We use this minibatch of $N$ generated examples to compute the gradient of the
minibatch log-likelihood $\LL$, where
\begin{align*}
    \LL(\TT_c, \TT_r, \TT_o) = 
    \frac{1}{N} \sum_{n=1}^N \sum_{t=k+1}^{T} \log p(\vz_t^n = \vz_t^{*,n}| 
    \vz_{<t}^{*,n}, \vx_{\leq t}^n).
\end{align*}
As this is an anonymous object tracking system, the model is given the
ground-truth locations of the object for the first $k$ frames.

Another training criterion is possible, if our prediction $\tilde{\vz}_t$ as
each step $t$ is a differentiable function. In this case, we let the model
freely track an object given a training video sequence and maximize the
log-probability of the ground-truth location {\em only} at the last frame:
\begin{align}
    \label{eq:free_cost}
    \LL(\TT_c, \TT_r, \TT_o) = 
    \frac{1}{N} \sum_{n=1}^N \log p(\vz_T^n = \vz_T^{*,n} | \tilde{\vz}_{<T}^n,
    \vx_{\leq T}^n).
\end{align}
According to our preliminary experiments, we use this second strategy throughout
this paper.

Of course, in this case, there is no guarantee that any intermediate prediction
made by the model correspond to the correct object location. To avoid this
issue, we add the following auxiliary cost to the new cost above:
\begin{align}
    \label{eq:location_cost}
    \tilde{\LL}(\TT_c, \TT_r, \TT_o) =
    \frac{1}{N} \sum_{n=1}^N \sum_{t=k+1}^{T} \log p(\vz_t^n = \vz_t^{*,n} |
    \tilde{\vz}_{<t}^n, \vx_{\leq t}^n).
\end{align}
Minimizing this auxiliary cost encourages the model to following
the object in the intermediate frames.

In our case, the model predicts two points $\vz_t = \left[ x_0, y_0, x_1, y_0
\right]$ in the input frame which corresponds to the top-left $(x_0, y_0)$ and
bottom-right $(x_1, y_1)$ corners of a bounding box. We use a Gaussian
distribution with an fixed, identity covariance matrix, whose mean is computed
from $\vh_t$ (see Eq.~\eqref{eq:rnn_update}.) In order to reduce variance, we do
not sample from this distribution, but simply take the mean as the prediction:
\begin{align*}
    \tilde{\vz}_t = \E\left[ \vz_{t} | \tilde{\vz}_{<T}, \vx_{\leq T}\right].
\end{align*}
This effectively reduces the auxiliary cost in Eq.~\eqref{eq:location_cost} as
well as the main cost in Eq.~\eqref{eq:free_cost} to a mean-squared error.

In the case of using the modified selective attention model to preprocess the
input frame, there are two additional output elements which are the standard
deviation $\sigma$ and the stride $\delta$. As we consider both of these as real
values, this simply makes the predictive distribution to be a six-dimensional
vector, i.e., $\dim(\vz_t)=4$.

\subsection{Characteristics}

There are three main characteristics that set the proposed approach apart
from the previous works on visual object tracking. 

First, the proposed model is trained end-to-end, including object representation
extraction, object detection and object tracking. The model works directly on
the raw pixel intensities of each frame. This is unlike conventional object
tracking systems, in which appearance modeling is considered largely separate
from the actual tracking system~\citep[see, e.g.,][]{li2013survey}. This largely
prevents potential performance degradation from having suboptimal,
hand-engineered object representation and detector. 

Second, the proposed model works with anonymous objects by design. As we train a
model with a large set of generic-shaped objects offline, the model learns to
detect a generic object that was pointed out initially rather than to detect a
certain, predefined set of objects. As the proposed model is a recurrent neural
network which can maintain the history of the object's trajectory, it implicitly
learns to find a region in an input frame which has a similar activation pattern
from the previous frames. In fact, human babies are known to be able to track
objects even before having an ability to classify it into one of the object
categories~\citep{ruff2001attention}.

Lastly, training is done fully off-line. We note first that this is both an
advantage and disadvantage of the proposed approach. This off-line training
strategy is advantageous in test time, as there is no need to re-tune any part
of the system. As the model parameters are fixed during test time, it will be
more straightforward to implement the trained model on a hardware, achieving a
desirable level of power consumption and speed~\citep{farabet2011large}. On the
other hand, this implies that the proposed system lacks the adaptability to
novel objects that are novel w.r.t. the shapes used during training, which is
exactly a property fully exploited by any existing tracking-by-detection visual
tracking system.  

\subsection{Related Work}
\label{sec:related}

As we were preparing this work, \citet{kahou2015ratm} independently proposed a
similar visual object tracker based on a recurrent neural network. Here let us
describe the similarities and differences between their recurrent attentive
tracking model (RATM) with the proposed tracking approach.

A major common feature between these two approaches is that both of these use a
recurrent neural network as a main component. A major difference between the
RATM and the proposed approach is in training. Both the RATM and the model
proposed in this paper use the intermediate locations of an object as an
auxiliary target (see Eq.~\eqref{eq:location_cost}.) \citet{kahou2015ratm}
report that this use of auxiliary cost stabilized the tracking quality, which is
further confirmed by our experiments presented later in this paper.  

A major difference is that \citet{kahou2015ratm} used a classification error,
averaged over all the frames, as a final cost, while we propose to use the final
localization error.  Furthermore, they use the selective attention mechanism
from \citep{gregor2015draw}, allowing the RATM only a small sub-region of the
whole canvas at each frame. This is contrary to the recurrent visual tracker
proposed in this paper which has access to the full frame every time.

Earlier, \citet{denil2012learning} proposed a visual object tracking system
based on deep neural networks. Their model can be considered as an intermediate
step away from the conventional tracking approaches toward the one proposed here
and by \citet{kahou2015ratm}. They used a restricted Boltzmann machine
\citep{Smolensky1986} as an object detection model together in a filtering-based
visual tracking (state-space model with particle filtering for inference.) 

Although they are not specifically for visual object tracking, two recent works
on image/video description generation tasks have shown that a recurrent network,
together with a convolutional network, is able to attend (detect and localize)
to objects in both an image and a video clip
\citep{xu2015show,yao2015describing}. Similarly, 
\citet{mnih2014ram} and \citet{ba2014multiple} showed that a recurrent network
tracks an object if it were trained to classify an object, or multiple objects,
in an image.

\section{Data Generation}
\label{sec:data}

We evaluate the proposed approach of visual anonymous object tracking on
artificially generated datasets. We vary the configurations of generating these
datasets to make sure that the following empirical evaluations support our claim
and conjectures about the proposed model. 

All the datasets are based on the cluttered MNIST.\footnote{
    \url{https://github.com/deepmind/mnist-cluttered}
}
Each video sequence used for training consists of 20 frames, and each frame is
$100 \times 100$ large. The cluttered MNIST was chosen as a basis of generating
further datasets, as one of the most important criterion we aim to check with
the proposed approach is the robustness to noise.  In order to make sure that
these clutters acts as both background noise and as objects hindering the sight
of the models, we put some clutters in a background layer and the others in a
foreground layer (overshadowing the target object.) Furthermore, the clutters
move rather than stay in the initial positions to make it more realistic.

The target has a random initial velocity $(v_{x_0}, v_{y_0})$ and position
$(x_0, y_0)$.  At each time frame, the position is changed by $(\Delta x_t,
\Delta y_t) = (kv_{x,{t-1}}, kv_{y,{t-1}})$ where $k$ is a hyper-parameter (0.1
in our experiments) correlated to the frame rate.  We change the velocity by
$(\Delta v_{x,t}, \Delta v_{y,t}) \sim \mathcal{N}(0, v'I)$, where $v'=0.1$ is a
hyper-parameter controlling the scale of velocity changes. This change in the
velocity introduces acceleration, making it more difficult for a tracking
system.

To ensure that our dataset is as realistic as possible, we include other
transformations. For example, at each time step, the target changes its scale by
a random factor $f=p \exp(\tilde{f})$, where $\tilde{f}\sim\mathcal{U}(-0.5,
0.5)$ and $p=0.1$ controls the magnitude of scale change. Finally,
the intensities of each clutter and the moving MNIST digit are uniform-randomly
scaled between 64 and 255 (before normalization.)

\paragraph{Multiple Digits}

We evaluate our model on two different cases. In the first case, there is only a
single digit moving around in each video sequence. Although there are clutters
in the background/foreground, this setting is sufficiently easy and can be
considered as a sanity check. We call this dataset {\bf MNIST-Single-Same}.

The second dataset, {\bf MNIST-Multi-Same}, contains frames of which each
contains more than one digits. More specifically, we generate each video
sequence such that there are two digits simultaneously moving around. In
order for the tracking system to work well, the system needs to be able to
detect the characteristics of the object in interest from the first few frames
(when the ground-truth locations are given) and maintain it throughout the
sequence.

\paragraph{Novel Test Digit Classes}

As our goal is to build an {\em anonymous} object tracking system. We evaluate a
trained model with sequences containing objects that do not appear during
training time. More specifically, we test the models on two sets of sequences
containing one or two MNIST-2 digits, where one MNIST-2 digit is created
by randomly overlapping two randomly selected normal MNIST digits on top of each
other \citep{wang2014}. We call these datasets {\bf MNIST-Single-Diff} and {\bf
MNIST-Multi-Diff}, respectively.

\paragraph{Generalization to Longer Video Sequence}

In all the cases, we evaluate a trained model on test sequences that are longer
than the training sequences. This is a necessary check for any model based on
recurrent neural networks, as some recent findings suggest that on certain tasks
recurrent neural networks fail to generalize to longer test
sequences~\citep{joulin2015inferring,grefenstette2015learning}. We vary the
lengths of test sequences among $\left\{ 20, 40, 80, 160 \right\}$, while all
the models are trained with 20-frame-long training sequences.

\section{Models and Training}
\label{sec:model_train}

We test five models on each of the four cases, {\bf  MNIST-\{Single,Multi\}-\{Same,Diff\}}.

\paragraph{Recurrent Visual Object Tracker (RecTracker-X)}

The first model, {\bf RecTracker-ID}, is the proposed recurrent visual object
tracker (see Sec.~\ref{sec:description}.) 
%As described earlier, it consists of a
%convolutional network and a recurrent network. In this case, we use an identity
%function to preprocess each input frame $\vx_t$:
%\begin{align}
%    \label{eq:preprocess_id}
%    m(\vx_t, \tilde{\vz}_{t-1}) = \vx_t,
%\end{align}
%where $\tilde{\vz}_{t-1}$ is ignored. This is equivalent to simply letting the
The tracker has a full, unadjusted view of the whole frame. Alternatively, the tracker
with attentive weight scheme imposes weighting mask of $N \times N$ Gaussian filters. 
In this paper, We evaluate {\bf RecTracker-Att-1} and {\bf RecTracker-Att-3}, where
$N$ is $1$ and $3$ respectively.

\paragraph{Convolutional Network Only Tracker (ConvTracker)}

The third model is a simpler variant of the proposed model where the recurrent
network is omitted from the proposed recurrent visual tracking model. Instead,
this model considers four frames (three preceding frames + current frame) and
predict the location of an object at the current frame. We only test the
identity preprocessing routine in Eq.~\eqref{eq:preprocess_id}.  We call this
model {\bf ConvTracker}.  Other than the omission of the recurrent network, all
the other details are identical to those of RecTracker-ID, RecTracker-Att-1 and
RecTracker-Att-3. 

\paragraph{Kernelized Correlation Filters based Tracker (KerCorrTracker)}

Lastly, we use the kernelized correlation filters-based visual tracker proposed
by \citet{henriques2015tracking}.\footnote{
    \url{http://home.isr.uc.pt/~henriques/circulant/}
}
This is one of the state-of-the-art visual
object tracking systems and follows a tracking-by-detection strategy (see
Sec.~\ref{sec:background}.) This third model is chosen to highlight the
differences between the existing tracking approaches and the proposed one.

%\paragraph{Convolutional Network}
%
%The convolutional network we use has the following structure. The first layer is
%a convolutional layer without pooling, the second and the third layer are both
%fully-connected layers, with the last layer emitting the target bounding box.  
%%\begin{itemize}
%%    \itemsep 0em
%%    \item \alert{The first layer is a convolutional layer, without pooling layer [Shall we present the experiment settings, i.e. the network
%%    	architecture hyperparameters here?  I feel that the numbers should
%%    	be instead presented in something like Experiment Settings section]}
%%    \item \alert{The second layer and the third layer are both fully-connected
%%    	layers, with the last layer emitting the target bounding box
%%    	coordinates (top-left, bottom-right corner)}
%%\end{itemize}

\subsection{Network Architectures}

We describe the architectures of each network--convolutional and recurrent
networks-- in Appendix~\ref{apx:architecture}.

\subsection{Training}

We describe the architectures of each network--convolutional and recurrent
networks-- in Appendix~\ref{apx:training}.

\section{Result and Analysis}
\label{sec:result}

\subsection{Evaluation Metric}

We use the intersection-over-union (IOU) as an evaluation metric.  The IOU is
defined as
\begin{align*}
    \IOU(\vz_t^*, \tilde{\vz}_t) = \frac{\left| M^* \cap \tilde{M} \right|}
    {\left| M^* \cup \tilde{M} \right|},
\end{align*}
where $M^*$ and $\tilde{M}$ are binary masks whose pixels inside a bounding box
(either ground-truth $*$ or predicted $\tilde{}$) are 1 and otherwise 0. A
higher IOU implies better tracking quality, and it is bounded between $0$ and
$1$.  For each video sequence, we compute the average IOU across all the frames.

\subsection{Quantitative Analysis}

\begin{table}[ht]
    \centering
    \tiny
    \begin{tabular}{l || c c c c | c c c c}
        & \multicolumn{4}{c|}{ConvTracker} & \multicolumn{4}{c}{RecTracker-ID} \\
        \hline 
        {Test Seq. Length} & 20 & 40 & 80 & 160 & 20 & 40 & 80 & 160 \\
        \hline 
        \hline 
        MNIST-Single-Same & $0.36\pm 0.12$ & $0.35\pm 0.11$ & $0.33\pm 0.10$ & $0.29\pm 0.10$ & $0.61\pm 0.11$ & $0.61\pm 0.11$ & $0.58\pm 0.12$ & $0.53\pm 0.13$ \\
        MNIST-Single-Diff & $0.37\pm 0.13$ & $0.35\pm 0.11$ & $0.33\pm 0.10$ & $0.29\pm 0.10$ & $0.48\pm 0.08$ & $0.49\pm 0.09$ & $0.48\pm 0.08$ & $0.46\pm 0.10$ \\
        \hline
        MNIST-Multi-Same & $0.17\pm 0.11$ & $0.15\pm 0.09$ & $0.14\pm 0.08$ & $0.13\pm 0.08$ & $0.36\pm 0.20$ & $0.31\pm 0.19$ & $0.28\pm 0.18$ & $0.26\pm 0.17$ \\
        MNIST-Multi-Diff & $0.17\pm 0.13$ & $0.15\pm 0.11$ & $0.14\pm 0.10$ & $0.13\pm 0.10$ & $0.33\pm 0.14$ & $0.29\pm 0.14$ & $0.26\pm 0.13$ & $0.24\pm 0.12$ \\
    \end{tabular}
    \caption{Average and standard deviation of IOU's computed over 500 test sequences using the
        ConvTracker and RecTracker-ID with different datasets and different
        lengths of test sequences. From this table, it is clear that the
        RecTracker-ID performs better than the ConvTracker which does not have a
    recurrent network unlike the RecTracker-ID.}
    \label{tab:conv_vs_rec}
\end{table}

\paragraph{Importance of Recurrent Network}

First, we compare the {\bf ConvTracker} and {\bf RecTracker-ID}. They both use
the identity preprocessing routine, and the only difference is that the latter
has a recurrent network while the former does not. The results, in terms of the
average IOU, are presented in Table~\ref{tab:conv_vs_rec}.

In this table, the importance of having a recurrent layer is clearly
demonstrated. Across all the cases--different data configurations and test
sequence lengths-- the RecTracker-ID significantly outperforms the ConvTracker.
Also, we notice that the tracking quality degrades as the length of test
sequences increases (up to 8 folds.) 

\begin{table}[ht]
    \centering
    \tiny
    \begin{tabular}{l || c c c c | c c c c}
        & \multicolumn{4}{c|}{RecTracker-ID} &
        \multicolumn{4}{c}{RecTracker-Att-$1$} \\
        \hline 
        {Test Seq. Length} & 20 & 40 & 80 & 160 & 20 & 40 & 80 & 160 \\
        \hline 
        \hline 
        MNIST-Single-Same & $0.61\pm 0.11$ & $0.61\pm 0.11$ & $0.58\pm 0.12$ & $0.53\pm 0.13$ & $0.59\pm 0.14$ & $0.58\pm 0.14$ & $0.54\pm 0.14$ & $0.48\pm 0.15$  \\
        MNIST-Single-Diff & $0.48\pm 0.08$ & $0.49\pm 0.09$ & $0.48\pm 0.08$ & $0.46\pm 0.10$ & $0.64\pm 0.06$ & $0.64\pm 0.06$ & $0.61\pm 0.07$ & $0.56\pm 0.09$  \\
        \hline 
        MNIST-Multi-Same & $0.36\pm 0.20$ & $0.31\pm 0.19$ & $0.28\pm 0.18$ & $0.26\pm 0.17$ &  $0.37\pm 0.22$ & $0.35\pm 0.22$ & $0.29\pm 0.22$ & $0.25\pm 0.20$ \\
        MNIST-Multi-Diff & $0.33\pm 0.14$ & $0.29\pm 0.14$ & $0.26\pm 0.13$ & $0.24\pm 0.12$ &  $0.41\pm 0.19$ & $0.35\pm 0.18$ & $0.31\pm 0.17$ & $0.28\pm 0.16$ \\
    \end{tabular}
    \caption{Average and standard deviation of IOU's computed over 500 test sequences using the
        RecTracker-ID and RecTracker-Att-$1$. The results by {\bf
    RecTracker-ID} are identical to those in Table~\ref{tab:conv_vs_rec}.}
        \label{tab:att1_vs_id}
\end{table}

\paragraph{Effect of Attentive Weight Scheme}

Next, we evaluate the effect of the attentive weight scheme (see
Eq.~\eqref{eq:preprocess_att} and surrounding text) against the simple
identity preprocessing scheme (see Eq.~\eqref{eq:preprocess_id} and surrounding
text.) From Table~\ref{tab:att1_vs_id}, we notice an interesting pattern. When
the training and test shapes are of the same classes (similar shapes), it is
better to use the identity preprocessing scheme ({\bf RecTracker-ID}). On the
other hand, {\bf RecTracker-Att-$1$} outperforms the simpler one, when the
shapes are different between training and test.

\begin{table}[h]
	\centering
	\tiny
	\begin{tabular}{l || c c c c | c c c c}
		& \multicolumn{4}{c|}{RecTracker-Att-$1$} &
		\multicolumn{4}{c}{RecTracker-Att-$3$} \\
		\hline 
		{Test Seq. Length} & 20 & 40 & 80 & 160 & 20 & 40 & 80 & 160 \\
		\hline 
		\hline 
		MNIST-Single-Same & $0.59\pm 0.14$ & $0.58\pm 0.14$ & $0.54\pm 0.14$ & $0.48\pm 0.15$  & $0.57\pm 0.14$ & $0.55\pm 0.14$ & $0.53\pm 0.15$ & $0.47\pm 0.15$ \\
		\hline
		MNIST-Multi-Same & $0.37\pm 0.22$ & $0.35\pm 0.22$ & $0.29\pm 0.22$ & $0.25\pm 0.20$ & $0.36\pm 0.22$ & $0.32\pm 0.21$ & $0.27\pm 0.21$ & $0.24\pm 0.19$ \\
		\hline
		MNIST-Single-Diff & $0.64\pm 0.06$ & $0.64\pm 0.06$ & $0.61\pm 0.07$ & $0.56\pm 0.09$ & $0.61\pm 0.06$ & $0.61\pm 0.06$ & $0.59\pm 0.06$ & $0.56\pm 0.10$ \\
		\hline
		MNIST-Multi-Diff & $0.41\pm 0.19$ & $0.35\pm 0.18$ & $0.31\pm 0.17$ & $0.28\pm 0.16$ & $0.38\pm 0.17$ & $0.33\pm 0.17$ & $0.30\pm 0.16$ & $0.27\pm 0.16$ \\
	\end{tabular}
	\caption{Average and standard deviation of IOU's computed over 500 test sequences using the
		RecTracker-Att-$1$ and RecTracker-Att-$3$}
	\label{tab:1_vs_3}
\end{table}

\begin{figure}
	\begin{minipage}{0.48\textwidth}
		\centering
		\includegraphics[width=\columnwidth]{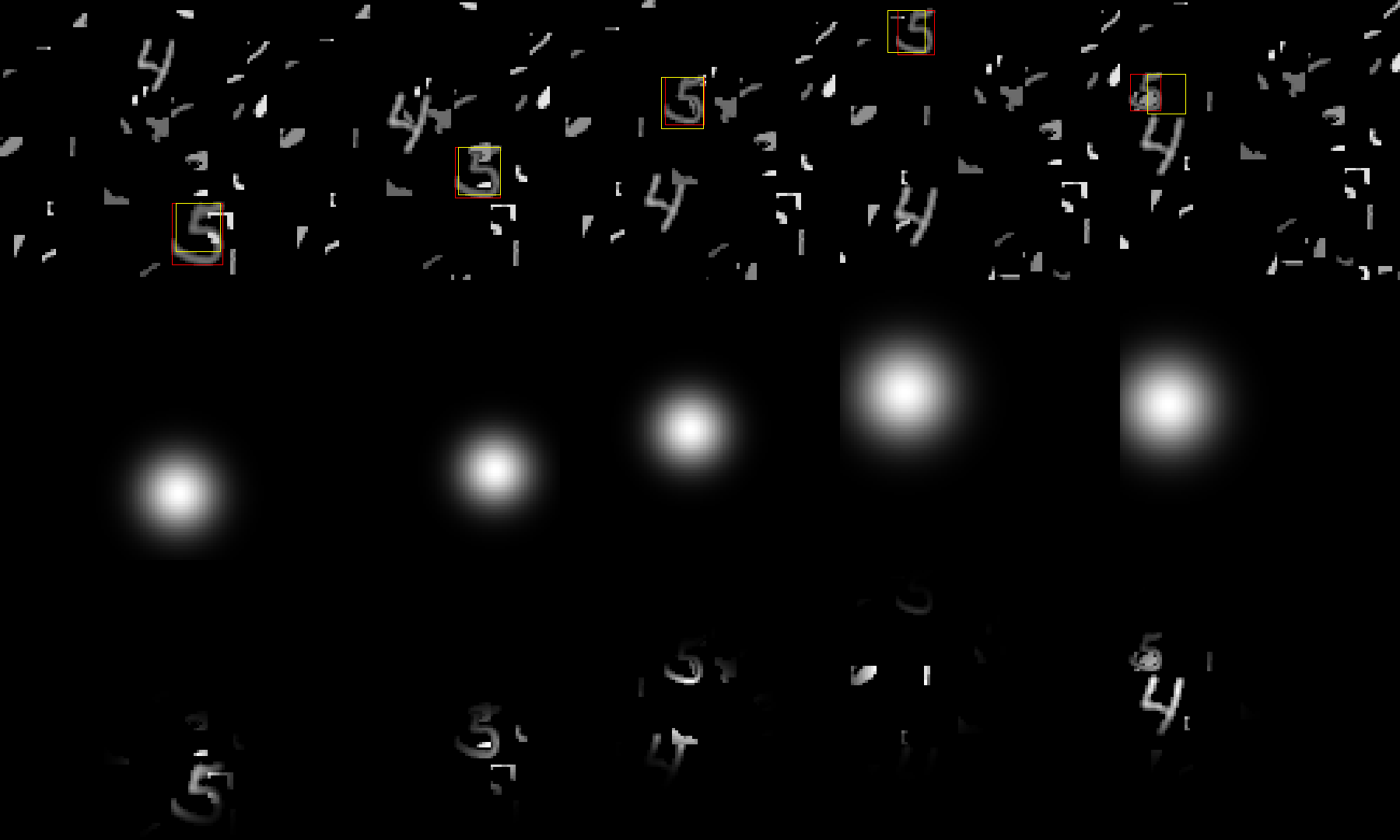}
		
		(a)
	\end{minipage}%
	\hfill
	\begin{minipage}{0.48\textwidth}
		\centering
		\includegraphics[width=\columnwidth]{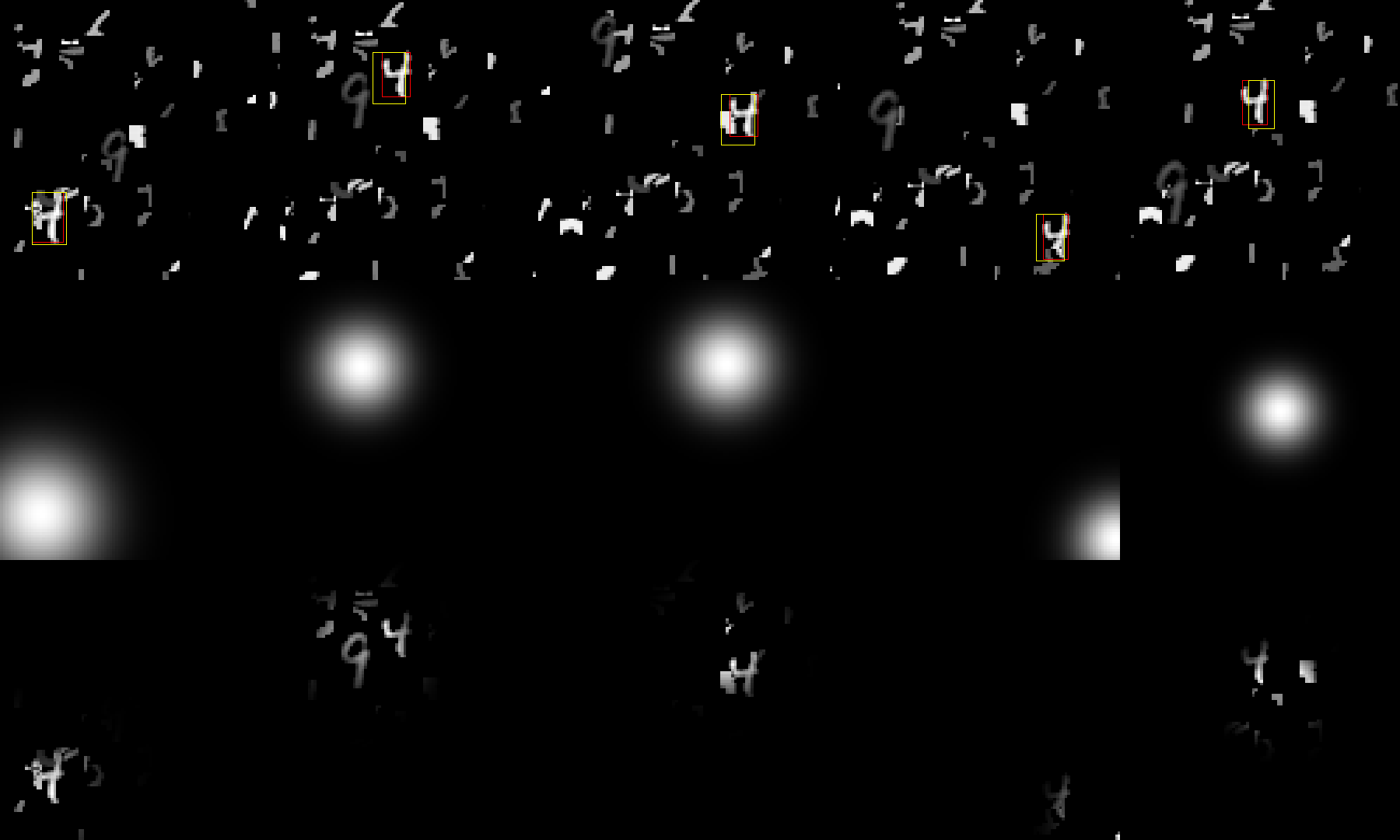}
		
		(b)
	\end{minipage}
	\caption{(a) RecTracker-Att-1 and (b) RecTracker-Att-3.  Image sequence, label and prediction shown at the top row, while the masks are shown at middle, and the masked images are shown at bottom.  The mask patterns look alike between models.}
	\label{fig:att-1-and-3}
\end{figure}

\paragraph{Comparison between different $N$}

We then compare \textbf{RecTracker-Att-$1$} with \textbf{RecTracker-Att-$3$}.
As shown in Table~\ref{tab:1_vs_3}, the performance between the two models
did not differ that much.  As it turns out, as long as the center Gaussian filter focuses
on the target, the model has not motivation to adjust the other ones. Manual 
inspection reveal these ``outer'' filters are often placed outside the canvas. 
%The reason is that the attention pattern of the
%latter model, which is a set of Gaussian distributions with centers
%aligned to a $3\times 3$ grid, only extends the $N=1$ case by adding a set of
%"outer" filters.  If the center Gaussian is focused on the object, the outer
%Gaussians essentially will not help much for prediction since they are focused on either
%clutters, background, or even somewhere out of the canvas.  
In Fig.~\ref{fig:att-1-and-3}, we show the attention weights computed by
\textbf{RecTracker-Att-1} and \textbf{RecTracker-Att-1}.  Evidently, only one Gaussian 
is being used. 
%We can see that
%only one Gaussian is used, and other Gaussians are not visible since they're
%out of the canvas.
%Figure~\ref{fig:att-1-and-3} further revealed that the
%\textbf{RecTracker-Att-$3$} learned a very similar attention pattern to
%\textbf{RecTracker-Att-$1$}, which demonstrates that the outer Gaussians did
%little help.

\begin{table}[h]
    \centering
    \tiny
    \begin{tabular}{l || c c c c | c c c c}
        & \multicolumn{4}{c|}{RecTracker-Att-$1$} &
        \multicolumn{4}{c}{KerCorrTracker} \\
        \hline 
        {Test Seq. Length} & 20 & 40 & 80 & 160 & 20 & 40 & 80 & 160 \\
        \hline 
        \hline 
        MNIST-Single-Same & $0.59\pm 0.14$ & $0.58\pm 0.14$ & $0.54\pm 0.14$ & $0.48\pm 0.15$  & $0.37\pm 0.28$ & $0.29\pm 0.26$ & $0.20\pm 0.22$ & $0.13\pm 0.15$ \\
        \hline
        MNIST-Multi-Same & $0.37\pm 0.22$ & $0.35\pm 0.22$ & $0.29\pm 0.22$ & $0.25\pm 0.20$ & $0.41\pm 0.26$ & $0.32\pm 0.26$ & $0.24\pm 0.23$ & $0.16\pm 0.17$ \\
        \hline
        MNIST-Single-Diff & $0.64\pm 0.06$ & $0.64\pm 0.06$ & $0.61\pm 0.07$ & $0.56\pm 0.09$ & $0.54\pm 0.27$ & $0.47\pm 0.28$ & $0.36\pm 0.26$ & $0.25\pm 0.22$ \\
        \hline
        MNIST-Multi-Diff & $0.41\pm 0.19$ & $0.35\pm 0.18$ & $0.31\pm 0.17$ & $0.28\pm 0.16$ & $0.55\pm 0.27$ & $0.46\pm 0.27$ & $0.34\pm 0.24$ & $0.23\pm 0.20$ \\
    \end{tabular}
    \caption{Average and standard deviation of IOU's computed over 500 test sequences using the
        RecTracker-Att-$1$ and KerCorrTracker with MNIST-1 and MNIST-2 test sets and different lengths of test sequences.}
        \label{tab:rec_vs_ker}
\end{table}

%\alert{This is
%especially apparent when these models are evaluated on the case of multiple
%digits with longer test sequences.}

\paragraph{Comparison against Tracking-by-Detection}

Table~\ref{tab:rec_vs_ker} compares the tracking quality between the proposed
recurrent tracking model {\bf RecTracker-Att-1} and the kernelized correlation
filters based one {\bf KerCorrTracker} by \citet{henriques2015tracking}. We
observe that {\bf RecTracker-Att-1} outperforms the {\bf KerCorrTracker} when
there's only a single object in a test sequence. However, when there are two
objects in a sequence and the model was asked to track only one of them, the
{\bf RecTracker-Att-1} and {\bf KerCorrTracker} perform comparably, but only do so with shorter sequences. Tracking-by-detection focuses
exclusively on local region and get less distracted. However, longer sequences
have higher probability of distraction (by either clutter or another digit). Consequently,
retaining longer history allows our model to steer towards our target better.
Finally, one noticeable difference is that standard deviation of {\bf RecTracker-Att-1} is 
one order of magnitude smaller, indicating a much tighter tracking.
Putting these results together, we note the followings. The under-performing {\bf
ConvTracker}  clearly demonstrates that it is important for a tracking model to
be capable of capturing temporal dynamics. Among the proposed recurrent tracking
models, the {\bf RecTracker-ID} performs better when the same types of objects
are used both during training and test, but in the other case the {\bf
RecTracker-Att-1} does better. This is similar to the observation we made when
these models were tracking a single object. 

We however remind readers that these results should be taken with a grain of salt.
It is unclear how the proposed tracking model can be extended to track
arbitrarily many objects. Also, a more accurate comparison between the proposed models
and tracking-by-detection needs to involve natural scenes, real objects and
dynamics. Under such scenarios, many extensions are called for in order to make
the model robust. We leave this as future research.

%It is however unclear how the proposed tracking model can be extended to track
%arbitrarily many objects, and thus, the multi-object tracking experiments are
%preliminary. In the future, the investigation is needed on the extension of the
%proposed recurrent tracking model to an arbitrary number of objects.

\subsection{Visualization of Tracking}

\begin{figure}[t]
\begin{minipage}{0.48\textwidth}
\centering
\includegraphics[width=\columnwidth]{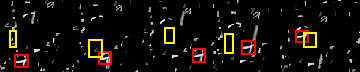}

(a)
\end{minipage}%
\hfill
\begin{minipage}{0.48\textwidth}
\centering
\includegraphics[width=\columnwidth]{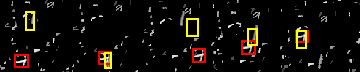}

(b)
\end{minipage}

\begin{minipage}{0.48\textwidth}
\centering
\includegraphics[width=\columnwidth]{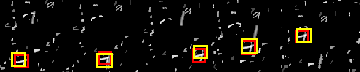}

(c)
\end{minipage}%
\hfill
\begin{minipage}{0.48\textwidth}
\centering
\includegraphics[width=\columnwidth]{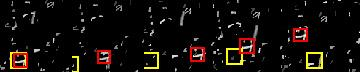}

(d)
\end{minipage}

\caption{(a) ConvTracker, (b) RecTracker-ID, (c) RecTracker-Att-1 and (d)
KerCorrTracker}
\label{fig:visualization}
\end{figure}

We have prepared video clips of tracking results by all the tested models at
\url{http://barclayii.github.io/tracking-with-rnn}. See Fig.~\ref{fig:visualization} for one such example.

Our visual inspection reveals that

\begin{enumerate}
    \item In most cases \textbf{ConvTracker} fail at tracking an anonymous
        object. This suggests the importance of having an explicit memory which
        is lacking from the \textbf{ConvTracker}. 

    \item Although the quantitative analysis based on IOU,
        \textbf{RecTracker-ID} almost always underperforms compared to
        \textbf{RecTracker-Att-$N$}'s, the qualitative analysis reveals that it
        still tracks an object fairly well. This suggests that the IOU may not
        be the optimal measure of tracking quality.

    \item \textbf{RecTracker-Att-$N$}'s often confuse the object being tracked,
        and this happens especially when another brighter object passes nearby.
        This may be due to the fact that the masking with Gaussian attentive is 
        too simplistic: it penalizes darker pixels more than brighter ones, making a
        brighter object stand out more. 
\end{enumerate}

\section{Conclusion}
\label{sec:conclusion}

In this paper, we proposed an end-to-end trainable visual object tracking model
based on convolutional and recurrent neural networks. Unlike conventional
tracking approaches, the full pipeline of visual tracking--object
representation, object extraction and location prediction-- is jointly tuned to
maximize the tracking quality. The proposed tracking model combines many recent
advances in deep learning and was found to well perform on challenging,
artificially generated sequences. 

We consider this work as a first step toward building a full end-to-end
trainable visual object tracking system. There are a number of issues to be
investigated and resolved in the future. First, the proposed models must be
evaluated on natural scenes with real objects and their dynamics. Second, there
needs to be research on algorithms to adapt a pre-trained model online. Third,
we need to find a network architecture that can track an arbitrary number of
objects without a predefined upper limit.

%\subsubsection*{Acknowledgments}
%
%\alert{TBD}

\bibliography{deeprectracker}
\bibliographystyle{iclr2016_conference}

\appendix

\section{Network Architectures}
\label{apx:architecture}

\paragraph{Convolutional Network}

We use a single convolutional layer with 32 $10 \times 10$ filters. These
filters are applied with stride 5 to the input frame. As it is important to
maintain as much spatial information as possible for tracking to work well, we
do not use any pooling. This convolutional layer is immediately followed by an
element-wise $\tanh$.

In the case of {\bf ConvTracker}, a fully-connected layer with 200 $\tanh$ units
follows the convolutional layer. This fully-connected layer also receives as the
input the predicted locations of the four preceding frames. 

%Finally, another fully connected
%regression layer emits the top-left and bottom-right coordinates as the model's
%predicted bounding box.  Both fully connected layers use hypertangent as
%activation function.

\paragraph{Recurrent Network}

We use 200 gated recurrent units~\cite[GRU,][]{cho2014learning} to build a
recurrent network. At each time step, the activation of the convolutional layer
(see above) and the predicted object location $\tilde{\vz}_{t-1}$ in the
previous frame are fed into the recurrent network.

%At each time step, the current image is passed through a
%convolutional net with only one layer, its configuration is identical to the
%first layer of {\bf ConvTracker}. The CNN feature maps, as well as the
%model's previously predicted bounding box, is then passed into the recurrent
%layer.  The GRU's output is then processed by a fully-connected regression layer
%to produce the bounding box, targeting the object in current frame.
%
%In {\bf RecTracker-Att-$N$}, another fully-connected regression layer is
%connected to the GRU's output to produce the parameters of attention.

%The ground truth bounding box is exposed to the network only for the first time
%frame, identifying the tracking target. On all subsequent frames the model
%solely relies on the current image and its own prediction.

\section{Training}
\label{apx:training}

We train each model up to 50 epochs, or until the training cost stops improving,
using a training set of 3,200,000 randomly-generated examples.  We use RMSProp,
which was implemented according to \citep{graves2013generating}, with
minibatches of size 32 examples.

\end{document}